\documentclass[11pt]{article}

\usepackage[preprint]{acl}

\usepackage{times}
\usepackage{latexsym}

\usepackage[T1]{fontenc}
\usepackage[utf8]{inputenc}

\usepackage{xspace}                 % handle spacing at the end of macros you define
\usepackage{float}

\usepackage{microtype}

\usepackage{inconsolata}

\usepackage{graphicx}

\title{\dataset: A Structured Evaluation Framework of Hierarchical Errors in Dialogue Summaries}

\author{
 \textbf{Sahana Ramnath\textsuperscript{1}}\thanks{Work done during an internship at Capital One.},
 \textbf{Nima Chitsazan\textsuperscript{2}},
 \textbf{Mingyang Zhou\textsuperscript{2}},
 \textbf{Chia-Hsuan Lee\textsuperscript{2}},
\\
 \textbf{Shi-Xiong Zhang\textsuperscript{2}},
 \textbf{Stephen Rawls\textsuperscript{2}},
 \textbf{Sambit Sahu\textsuperscript{2}},
 \textbf{Sangwoo Cho\textsuperscript{2}},
\\
 \textbf{Xiang Ren\textsuperscript{1}},
 \textbf{Genta Indra Winata\textsuperscript{2}},
 \textbf{Akshaj Kumar Veldanda\textsuperscript{2}},
\\
 \textsuperscript{1}University of Southern California,
 \textsuperscript{2}Capital One
\\
 \small{
   \textbf{Contact:} \href{mailto:sramnath@usc.edu}{sramnath@usc.edu}
 }
}

\newcommand{\taxonomy}{\textsc{Dial-SummEr}\xspace}
\newcommand{\dataset}{\textsc{Dial-SummEr}\xspace}
\newcommand{\levela}{\textsc{Dialogue-Level}\xspace}
\newcommand{\levelb}{\textsc{Within-Turn-Level}\xspace}

\newcommand{\anthropictest}{\textsc{Anthropic-Test}\xspace}

\newcommand{\gptfive}{\textsc{GPT-5}\xspace}
\newcommand{\claude}{\textsc{Claude-Sonnet-4.5}\xspace}
\newcommand{\llama}{\textsc{Llama}\xspace}
\newcommand{\qwen}{\textsc{Qwen}\xspace}

\newcommand{\llamasumm}{\textsc{Llama-3.3-70B-Instruct}\xspace}

\newcommand{\llamafull}{\textsc{Llama-3.1-70B-Instruct}\xspace}
\newcommand{\claudefull}{\texttt{claude-sonnet-4-5-20250929}\xspace}
\newcommand{\qwenfull}{\textsc{Qwen2.5-32B-Instruct}\xspace}
\newcommand{\turnseq}{\textsc{Wrong Turn Sequence}\xspace}
\newcommand{\missturn}{\textsc{Missed Turn}\xspace}
\newcommand{\spkmis}{\textsc{Speaker Misattribution}\xspace}
\newcommand{\spkbias}{\textsc{Speaker Identity Bias}\xspace}
\newcommand{\viewpt}{\textsc{Viewpoint Distortion}\xspace}
\newcommand{\wronglink}{\textsc{Wrong Linking}\xspace}
\newcommand{\chmean}{\textsc{Changed Meaning}\xspace}

\newcommand{\missconv}{\textsc{Missed Conversation}\xspace}
\newcommand{\extconv}{\textsc{Extrinsic Conversation}\xspace}
\newcommand{\extcont}{\textsc{Extrinsic Context}\xspace}
\newcommand{\hall}{\textsc{Hallucination}\xspace}

\usepackage{soul}
\sethlcolor{lblue}
\usepackage{makecell}
\usepackage{booktabs}
\usepackage{dsfont}
\usepackage{fdsymbol}
\usepackage{multirow}
\usepackage{colortbl}
\usepackage{longtable}

\definecolor{MyViolet}{rgb}{0.5, 0.0, 1.0}
\definecolor{lightpurpletwo}{rgb}{0.706, 0.655, 0.839}
\definecolor{lightbluetwo}{rgb}{0.62, 0.77, 0.91}

\colorlet{lblue}{lightbluetwo!100}
\colorlet{lviolet}{lightpurpletwo!100}
\colorlet{lgreen}{green!35}
\colorlet{lyellow}{yellow!15}

\begin{document}
\maketitle

\begin{abstract}

Dialogues are a predominant mode of communication for humans, and it is immensely helpful to have automatically generated summaries of them (e.g., to revise key points discussed in a meeting, to review conversations between customer agents and product users). 
Prior works on dialogue summary evaluation largely ignore the complexities specific to this task:
(i) shift in \textit{\textbf{structure}}, from multiple speakers discussing information in a scattered fashion across several turns, to a summary's sentences, and (ii) shift in \textit{\textbf{narration viewpoint}}, from speakers' first/second-person narration, standardized third-person narration in the summary. In this work, we introduce our framework \textbf{\taxonomy} to address the above. We propose \taxonomy's taxonomy of errors to comprehensively evaluate dialogue summaries at two hierarchical levels: \colorbox{lblue}{\textbf{\levela}} that focuses on the broader speakers/turns, and \colorbox{lviolet}{\textbf{\levelb}} that focuses on the information talked about inside a turn. 
We then present \taxonomy's dataset composed of dialogue summaries manually annotated  with our taxonomy's fine-grained errors. 
% Our framework's purpose is two-fold: to observe common dialogue summarization errors and trends, and to further experiment if LLM-Judges are capable of detecting these errors. 
We conduct empirical analyses of these annotated errors, and observe interesting trends (e.g., turns occurring in \textit{middle} of the dialogue are the most frequently missed in the summary, extrinsic hallucinations largely occur at the \textit{end} of the summary). We also conduct experiments on  LLM-Judges' capability at detecting these errors, through which we demonstrate the challenging nature of our dataset, the robustness of our taxonomy, and the need for future work in this field to enhance LLMs' performance in the same. Code and inference dataset coming soon. %We submit our inference dataset and all codes as supplementary material.

% Following this proposed taxonomy, we develop a dataset of dialogue summaries and human annotated fine-grained errors of the same using our taxonomy; we release this dataset 
% we annotate an evaluation dataset
% with different categories of errors in the generated dialogue summary from the Large language model.

\end{abstract}

\section{Introduction} \label{sec:intro}
\begin{figure*}
    \centering
    \includegraphics[width=0.95\textwidth]{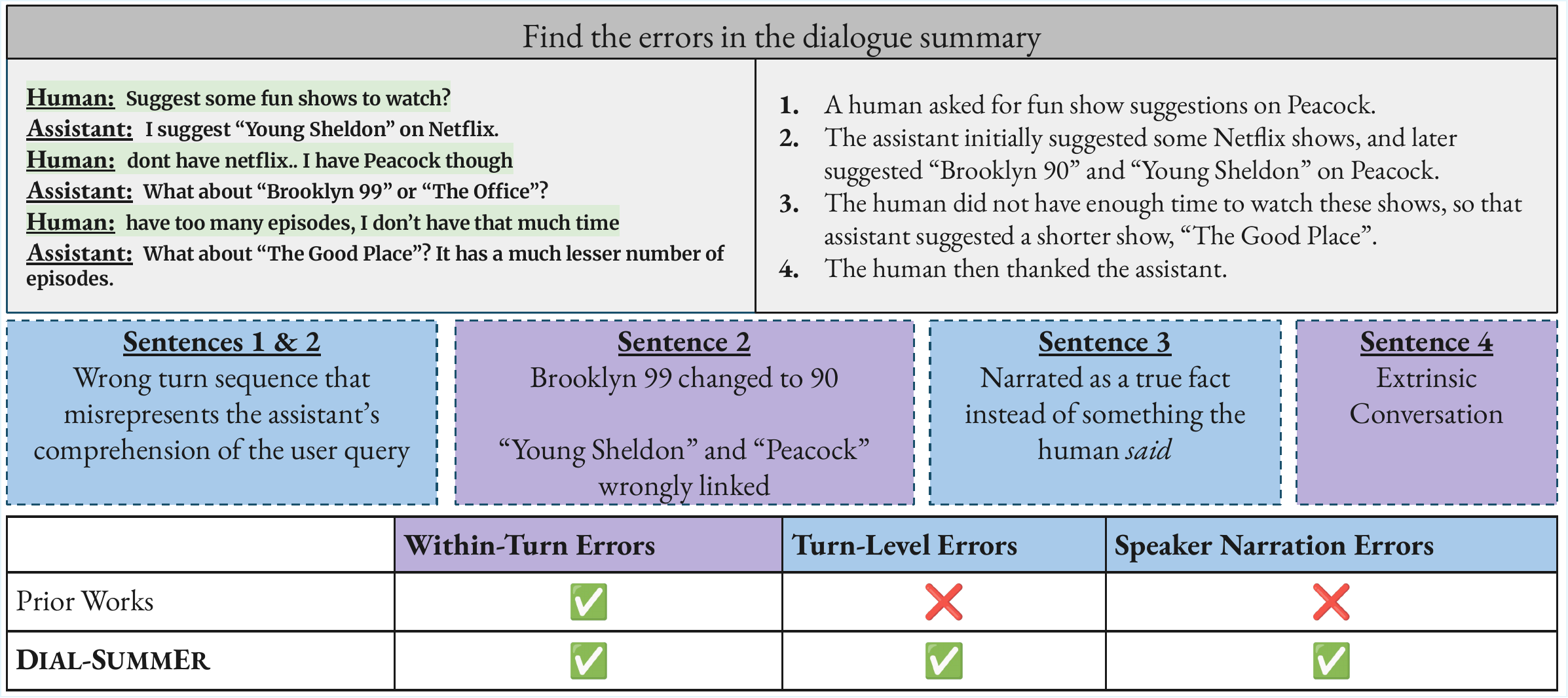}
    \caption{Our framework \textbf{\taxonomy} addresses errors that arise due to the shift in \textit{structure} and \textit{narration viewpoint} from a dialogue to its summary. Our framework is composed of (i) an error taxonomy with both dialogue-level and within-turn-level errors, (ii) a human-annotated dialogue-summary inference dataset with the same.}
    \label{fig:new-intro-fig}
\end{figure*}
% \begin{figure}
%     \centering
%     \includegraphics[width=0.95\linewidth]{images/ACL Intro Figure_cropped.pdf}
%     \caption{\textbf{\taxonomy} addresses the shift in structure and narration faced during dialogue summarization.\sahana{make a mapping other taxonomies versus ours}}
%     \label{fig:intro-fig}
% \end{figure}
% 1. Dialogue summarization is important + some examples + citations 
% 2. in our work, we provide a new taxonomy that includes specific look at the speakers and the turns (customized to dialogue summaries)
% 3. Then we annotate and release a dataset for these categories
% 4. Lastly, we show preliminary LLM as a judge experiments on this dataset

Dialogue has always been an ubiquitous part of every person's life, even more so today, with the widespread technological advancements that allow people to communicate with other people or with AI assistants through written conversations \cite{mcnichols2025studychat,jing-etal-2025-scale,zhenglmsys}. 
% Aside from personal conversations, dialogue interactions are heavily prevalent in domains such as education, travel/consumerism (for example, chats between a client and a human/AI customer agent), decision-making, medicine, etc. 
Automatically (LLM-)generated dialogue summaries are extremely helpful to capture the salient information discussed in the conversation \cite{zhang-etal-2021-exploratory-study}; e.g., students can use the summaries to quickly revise key concepts, consumer agencies can use the summaries to analyze client satisfaction and agent performance, doctors can use summaries of their conversations with patients to improve efficiency, etc. \cite{jain-etal-2023-summarize,mullick2024long,ben-abacha-etal-2023-overview}.
% For example, students who converse with AI assistants to learn new topics \cite{mcnichols2025studychat} would find a summary of their chat helpful for revising key points. Travel and consumer agencies that utilize  (human/AI) customer agents to chat with their clients \cite{jing-etal-2025-scale} would like a summary of these chats to analyze client satisfaction, agent performance, etc. Dialogue summarization is hard, owing to the 
% In today’s world, there exist many situations where humans interact with language models in the format of a dialogue, such as general conversations or student study sessions with API models such as ChatGPT, MetaAI, Gemini, Claude, etc., or user interactions with bank/travel/consumer-website chatbots regarding questions about the latter’s products (either via text or via audio calls which can later be transcribed into text) 
% \cite{that-paper-I-read}. In each of these scenarios, it is desirable to obtain \textit{summaries} of the conversation at the end for various purposes; for example, the human might desire to to have a succinct summary of everything they learnt from the conversation, the assistant might desire to create a memory of the conversation history and the human's preferences  to use as knowledge for future conversations, a consumer agency might desire a summary of the conversation between the human customer and their chatbot so that they can further assist the human caller, etc. 
In order to be useful, these summaries have to be faithful to the source conversation (i.e., free of hallucinations) and complete with respect to the downstream user's requirements; LLM-generated summaries however are typically error prone \cite{tang-etal-2024-tofueval}. 
% Hence, it is essential to have a comprehensive list of errors to evaluate these summaries, to understand the quality of the LLM-summarizers being used, and to further improve them.
% - this also means that we need good evaluation models which can accurately judge the summaries based on the above criteria.
% In this work, we focus on the \textit{evaluation} of these dialogue summaries in terms of \textit{hallucination} and \textit{incompleteness}. \\

% \noindent Prior works on using LLMs for summarization and its evaluation largely focus on 
% developing inference datasets that have human or LLM annotations of errors \cite{xx} and on developing judges (LLMs or reward models) \cite{xx} for the same. However, most of these works focus on 
% summarization of formal and structured source contexts such as news articles \cite{qi-etal-2025-evaluating,tang-etal-2024-tofueval}. 
There exists a large body of prior work on summarization evaluation: \citet{fabbri2020summeval,qi-etal-2025-evaluating} focus on news/science article summarization, \citet{tang-etal-2024-tofueval,song-etal-2024-finesure} focus on dialogue summarization, \citet{krichene-etal-2024-faithful} focuses on chart summarization, etc. Existing works on dialogue summarization mainly extend the evaluation framework from news summarization works, and miss to include the unique characteristics of dialogue. For example, previous works do not take into account the fact that dialogues have a unique hierarchical structure (at the speaker-turn level, and at the within-a-turn level); they also do not acknowledge the errors that could occur when the narration perspective switches from the first/second to third person. Further, some taxonomies also suffer from errors that have overlapping or unclear definitions: for example, in \citet{tang-etal-2024-tofueval} ``contradiction'' error is largely unclear; a summary sentence which contains any other error in the taxonomy (such as an entity error or a tense error) can also be considered to have a contradiction error.
% \mingyang{Missed the citation and example here}% (such as \cite{xx} who acknowledge it as a limitation).

% \sahana{at end of intro, make enumerate of contributions}

In this work, we propose \textbf{\taxonomy} (\textbf{D}ialogue-\textbf{Summ}ary \textbf{Er}rors), our framework that addresses the unique hierarchical nature of dialogue summarization via a \textbf{novel error taxonomy} and \textbf{evaluation dataset}. Our framework's purpose is three-fold: to release a comprehensive set of errors that can be applied and extended to any dialogue summary evaluation, to observe common dialogue summary errors and trends, and to further experiment if LLM-Judges are capable of detecting these errors.

Firstly, \taxonomy's taxonomy contains hierarchical errors that cover hallucination, incompleteness and narration perspective errors that occur in dialogue summaries. We propose five novel error categories at the speaker/turn level (including errors that are caused by the shift in narration perspective), and five error categories adapted from prior works for the information discussed within a conversational turn.
% we present speaker and turn level errors that are novel (todo: fix this) and within-turn level errors that we adapt and/or modify from relevant prior works.
Secondly, we present \taxonomy's dataset composed on dialogue summaries and human-annotated fine-grained errors.
% Secondly, we present \taxonomy's human-annotated dialogue summary error analysis dataset that we create based on our taxonomy.
We create this inference dataset from 192 \textbf{multi-turn dialogues} from \anthropictest \cite{bai2022training} and their \textbf{summaries}. Each dialogue-summary pair in our dataset contains the following: 1. the list of fine-grained errors exhibited by the summary; 2. the summary sentences where each error occurred; 3. the annotators' comments explaining the error annotation decision. We present empirical analyses on our dataset, such as the frequency of errors across and within dialogue summaries, and the average position of errors in summaries. We also test the capability of few-shot open-weight and proprietary LLM-Judges in detecting these errors. We observe that the LLMs perform only moderately well, demonstrating the challenging nature of \taxonomy's dataset; we also observe that their hallucination detection improves when our taxonomy is provided in the prompt. Overall, we develop and present:
\begin{enumerate}
    \item A hierarchical error taxonomy to evaluate dialogue summaries (Section \ref{sec:taxonomy})
    \item An inference dataset composed of dialogue summaries and human-annotated, fine-grained summary errors (Section \ref{sec:datasets})
    \item Empirical analyses (Section \ref{sec:dataset-analysis}) and LLM-Judge experiments on the inference dataset (Section \ref{sec:judge-exp}).
\end{enumerate}
We release our annotated inference dataset to motivate future work on dialogue summary evaluation, including more datasets, better LLM-Judges, and subsequently, error correction and better automatic summarization of dialogues.

% Lastly, we present several experimental results on how various LLM-judges perform with respect to each of these errors; we find that judges X are bad at Y and blah blah. We also find that adding our taxonomy to the LLM-judge prompt improves the performance of the judge in detecting the errors. 

\section{Related Work}
\label{sec:related-work}

% \sahana{change to present tense}
\textbf{Summarization Evaluation.} SummEval \cite{fabbri2020summeval} identified core relevant properties that a summary has to satisfy (coherence, consistency, fluency, relevance) and released corresponding human annotations on a subset of CNN/DailyMail summaries \cite{nallapati-etal-2016-abstractive}; these human annotations could be used to calculate the correlation with evaluation scores by popular summarization metrics such as ROUGE \cite{lin-2004-rouge}, BERTScore \cite{zhangbertscore}, etc. in order to identify the best evaluation scoring strategies. Eventually, works such as \citet{lee-etal-2024-unisumeval,song-etal-2024-finesure,min-etal-2025-towards} proposed further fine-grained categories to measure the quality of model-generated summaries; \citet{wu2023large} performs summarization evaluation with LLMs taking on multiple static and dynamic evaluation roles. \citet{song-etal-2025-learning,oh-etal-2025-learning} take this one step further to use LLM feedback with respect to these fine-grained categories to improve summarization models through reinforcement learning.

\textbf{Dialogue summarization evaluation.} Following the same steps as in SummEval, DialSummEval \cite{gao-wan-2022-dialsummeval} releases human annotations for dialogue summaries from SAMSum \cite{gliwa-etal-2019-samsum} on the categories of coherence, consistency, fluency and relevance, to test the correlation between human annotations and automatic metrics. Later, TofuEval \cite{tang-etal-2024-tofueval} released human annotations on dialogue summaries from MediaSum \cite{zhu2021mediasum} and MeetingBank \cite{hu2023meetingbank} with fine-grained error categories (such as reasoning, entity misreferencing, tense/modality, etc. errors) that could be used to analyze LLM-Judges; recently, \citet{jing-etal-2025-scale} released synthetic hallucinated general/topic-specific summaries of customer/travel agent call transcripts and annotations on fine-grained error categories (factual consistency, adjective usage, knowledge congruence, style alignment) for the same which can be used to test LLM-Judges. \textit{\textbf{In our work}}, we develop fine-grained error categories to prioritize explainability and informativity of evaluation. We propose a hierarchical error taxonomy to mirror the structure of dialogues, and we include speaker/turn-based errors, to fill the gap left by existing works on the same. We discuss in Appendix \ref{app:related-works-extended} how prior taxonomies face issues of unclear definitions of errors; we address these issues by adding macro-level errors in our taxonomy and ensuring that all of them are clearly defined and mutually exclusive.

% \sahana{Is this too big? We can keep it at this length if we fall slightly short of 8 pages (right now I think we have 7.5), otherwise, we can move some of this to appendix}

\section{\taxonomy's taxonomy} \label{sec:taxonomy}
\begin{table*}[ht]
    \centering
    \resizebox{0.95\textwidth}{!}{%
    \begin{tabular}{p{\textwidth}}
    \toprule
    \underline{\textbf{CONVERSATION}} \\
    \textbf{Human:} Search for a steak and mushroom recipe \\
    \textbf{Assistant:} Oh, I love recipes! Let me see what I can find. \\
    \textbf{Human:} Sounds great. \\
    \textbf{Assistant:} Looking up steak and mushroom recipe online, here's one on Chowhound: ``Short ribs, roasted mushrooms and miso, in a smoked salt crust, with roasted beets and potato salad. What's not to like?'' \\
    \textbf{Human:} Sounds very yummy. \\
    \textbf{Assistant:} Are you in the mood for a homemade steak and mushroom recipe? \\ \bottomrule
    \end{tabular}
    }
    \resizebox{0.95\textwidth}{!}{%
    \begin{tabular}{p{0.5\textwidth}|p{0.5\textwidth}}
    \textbf{Summary sentence(s)} & \textbf{Error explanation} \\ \midrule
    The assistant inquired if the human was in the mood for ... and the human replied that it sounded yummy. & \turnseq error, the last two turns have been swapped. \\ \midrule
    The assistant suggested a recipe of ... and the human found it to sound yummy. $<$end$>$ & \missturn since the last assistant turn was skipped in the summary. \\ \midrule 
    The \textcolor{red}{assistant found the recipe to sound very yummy}. & \spkmis error, since the human said the highlighted words. \\ \midrule
    The human said \textcolor{red}{she} found the recipe to be ... & \spkbias since the summary assumes the (gender) identity of the human speaker. \\ \midrule
    The assistant \textcolor{red}{began looking up a recipe} ... & \viewpt, the summarizer does not know if the assistant actually performed this action. \\ \midrule
    The assistant suggested a recipe from ... with \textcolor{red}{roasted potato and beet salad}. & \wronglink since the summary linked the wrong adjectives and nouns. \\ \midrule
    A human asked an LLM assistant to search for a \textcolor{red}{steak and tofu} recipe. & \chmean error because the summary changed `mushroom' to `tofu'. \\ \midrule 
    A human asked an LLM assistant to search for a recipe for steak and mushroom \textcolor{red}{with a side of greens}. & \extconv since the highlighted text is new conversation added by the summarizer. \\ \midrule
    A human asked an LLM assistant to search for a steak recipe, \textcolor{red}{probably because they were hungry}. & \extcont error since the summarizer added its own reasoning about the human's request. \\ \midrule
    The assistant suggested a recipe from ... with roasted beets. & \missconv since the summary has missed the `potato salad'. \\ \bottomrule
    \end{tabular}
    }
    \caption{Demonstration of errors from \taxonomy's taxonomy. This dialogue is taken from the \textsc{Anthropic-Train} set \cite{bai2022training}, and the summary errors (highlighted in \textcolor{red}{red}) are demonstrative.}
    \label{tab:taxonomy-example}
\end{table*}

In our framework \taxonomy, we propose a novel \textbf{taxonomy} designed to categorize errors in dialogue summaries across two hierarchical levels: \textbf{\colorbox{lblue}{\levela}} captures macro-level inconsistencies related to the structural integrity of the conversation  and the turns. \textbf{\colorbox{lviolet}{\levelb}} addresses the micro-level semantic discrepancies focusing on the content within a specific utterance. 
The errors are defined below (visualized in Table \ref{tab:taxonomy}, Appendix \ref{app:taxonomy-visual}).
% Speaker errors happen when the summary misattributes what was said in the dialogue to the wrong speaker, when it makes assumptions of the speaker's gender/race/etc. identity, and presents what was said by the speaker as a true fact instead of as narration. Turn-based errors happen when the summary misorders the turns of the dialogue, or when whole turns are missing in the summary.  \sahana{Move from table to here.}

\subsection{Descriptions of Errors}

 \textbf{\colorbox{lblue}{\turnseq}:} The dialogue's turns are presented in the wrong sequence in the summary.

 \textbf{\colorbox{lblue}{\missturn}:} The summary has missed to mention one or more conversational turns.

 \textbf{\colorbox{lblue}{\spkmis}:} The speakers are swapped/misreferenced in the summary.

 \textbf{\colorbox{lblue}{\spkbias}:} The summary makes assumptions about a speaker's identity (gender/race/etc.,)

 \textbf{\colorbox{lblue}{\viewpt}:} The summary presents what was \textit{said} in the conversation as a true fact. \textbf{Further explanation:} The summary is ideally expected to have each sentence narrated (in third person) from the perspective of the \textit{speakers}; this is to ensure that if the speakers said something inaccurate or false (for example, wrong world-knowledge, or false information about themselves or their actions), the summary does not propagate the error. A summary is hence said to have a \viewpt error if any of its sentences are written fully or partially from the point of view of a third-person narrator who is \textit{not} a speaker, who assumes that everything said by the speakers is correct (for example, world-knowledge) or did actually happen (for example, the action the speaker said they did or are doing or will do). 

% \paragraph{\colorbox{lviolet}{\levelb}} focuses on the lower, turn-level information that is discussed \textit{inside} a turn in the dialogue. 
% For example, if the summary exaggerates something a speaker said, or changes a question by the speaker into a declarative statement, if it adds new conversation by a speaker, or misses some details mentioned by a speaker, etc.
% The \levelb errors include any change in the details present in the dialogue, any new details which have been added, and any details that have been missed. %The errors are as follows:
%This includes errors where the summary \textit{contradicts} what was said in the source dialogue (..), errors where the summary \textit{adds} new information not mentioned in the source dialogue (..) and errors where the summary has missed information discussed in the dialogue. 

 \textbf{\colorbox{lviolet}{\wronglink}:} The summary presents information present in the conversation but links the wrong items together. 

 \textbf{\colorbox{lviolet}{\chmean}:} The summary has changed a word or a phrase from the conversation in a way that \textit{changes the meaning} of what the speakers said. This error includes: \textit{exaggeration} of words/events by the summary, \textit{change in sentence type} from the dialogue to the summary (e.g., question in the dialogue written as a statement in the summary), \textit{semantic misinterpretation} of ambiguous or idiomatic phrases or homonyms, as well as \textit{any other changed conversation} from the dialogue to the summary.

 \textbf{\colorbox{lviolet}{\extconv}:} The summary mentions that a speaker said something which was not said in the source conversation.

 \textbf{\colorbox{lviolet}{\extcont}:} The summary contains external explanations for something discussed in the conversation (for example the definition for an entity, the reason why a speaker said something, correcting what a speaker said using world knowledge, etc.).

 \textbf{\colorbox{lviolet}{\missconv}:} The summary has missed \textit{any} information mentioned in the dialogue; this error is \underline{extremely harsh}. %\textbf{Note:} This is the harshest possible incompleteness error.

We present an example annotation of a dialogue summary using \taxonomy in Table \ref{tab:taxonomy-example}. All \colorbox{lblue}{\levela} errors are novel categories proposed in this work; \colorbox{lviolet}{\levelb} categories are adapted and extended from prior works such as \citet{tang-etal-2024-tofueval,song-etal-2025-learning}. Further, in accordance with prior literature, we classify these errors (except \viewpt) into two broad categories: \textit{hallucination} and \textit{incompleteness}. \missturn and \missconv are incompleteness errors; every other error (apart from \viewpt) is a hallucination error. We exclude \viewpt from this classification since \viewpt is a subjective error; it is acceptable in low-stakes cases, but becomes an issue in high-stakes cases (such as, when the dialogue is between a client and a bank's customer agent).

\textbf{Note on incompleteness based errors:} We show two incompleteness errors, one on each level of information. While hallucination errors are definitive in terms of right and wrong, incompleteness errors are present in a gray area. A harsh evaluation is to expect every single detail in the conversation to be present in the summary. A less harsh but \textit{vague} strategy to check whether the summary includes all ``necessary'' details; however, the question of what is necessary to include in a summary is entirely user-dependent (or in the case of data annotation, annotator-dependent). In our work, we present two levels of incompleteness: \missturn and \missconv. The former error checks for incompleteness on a broad level, checking whether every single turn in the conversation (excluding greeting-only turns which \textit{we} deem as trivial) is mentioned in the summary; note that we do not ask for every detail in every turn to be included in the summary, only that the summary has to acknowledge the presence of the turn. The \missconv error on the other hand is extremely harsh, and we check for \textit{any detail} in the conversation (again, excluding greetings) that is missing in the summary; this is the harshest possible evaluation of incompleteness. We leave other moderate definitions of incompleteness to future/domain-specific work. 

\subsection{Addressing the complexities in dialogue summary evaluation}
% \textbf{How does \taxonomy address the two complexities in dialogue summarization?} 
Our taxonomy address the shift in structure from the dialogue to the summary by having hierarchical \colorbox{lblue}{\levela} and \colorbox{lviolet}{\levelb} errors that mirror the same. Further, it addresses the issue of the shift in narration by having speaker and narration specific errors (\spkmis, \spkbias, \viewpt), all of which only occur when the narration switches from the speakers' first/second perspectives to the summary's third-person perspective.

% We present all errors in \taxonomy in Table \ref{tab:taxonomy}. We introduce five novel \levela errors; we adapt/modify general summarization errors from several prior works \cite{xx,yy,zz} to present five \levelb errors. 

% \viewpt is a special case that falls into neither of these categories; it
% \mingyang{Maybe consider including a direct comparison table between our proposed taxonomy with some existing taxonomy defined in the previous literature to directly show the extension we include in the data} - \sahana{done in the intro on a broad level, I will add details in the appendix soon}

\section{\dataset Dataset} \label{sec:datasets}
% \sahana{add full stop in captions}

% \sahana{in top table, show which is challenging, which is not. we need to understand which fine-grained errors at failing }
In this section, we describe the creation of our framework's dialogue summary evaluation dataset. 
% we gave the annotators a rough annotation scheme and they annotated - then we the authors categorized them into our final taxonomy? no, just do taxonomy then dataset then initial results
% We wanted to pick a realistic human-AI chatbot scenario, so we pick \anthropictest \cite{bai2022training} as our dataset. 
% \subsection{Errors in Dialogue Summarization}
% In this work, we focus on conversations between two \textit{speakers}: a human and an AI assistant. The human and the assistant speak in alternating \textit{turns}, and the conversation is always initiated by the human and ended by the assistant. For each conversation, we have an LLM-generated summary. For a given conversation $C$ and its summary $S$, we use an LLM-judge $J$ to determine all the hallucination and incompleteness-related errors in $S$; we use the term \textit{instance} to refer to a $C,S$ pair. 
% \subsection{\dataset Annotation} \label{sec:dataset-annotation}
% \mingyang{Is there a reason why we pick this data to source the dialogues?} - \sahana{does this line sound satisfactory?}
We start with the aim of selecting a high-quality, \textit{multi-turn} dialogue dataset that has realistic, detailed (but not overly verbose or technical/niche) conversations; we select \anthropictest \cite{bai2022training}, a widely used dataset that contains human-AI conversations on various topics (e.g., casual chat, information-seeking, decision-making). We pick 200 conversations from \anthropictest that have four or more turns.\footnote{We take the ``chosen'' assistant response in the dataset.} We generate summaries for them using \llamasumm (prompt in Appendix \ref{app:all-prompts}) with greedy decoding: we instruct it to generate summaries that are  faithful to the dialogue, concise but still including all necessary details, and presented as a paragraph. 
For each conversation + summary pair, for each \taxonomy error, human annotators are instructed to mark ``YES'' if the summary exhibits that error, and ``NO'' if not; When they capture a certain error, they are also instructed to write down the reason for their decision on the detected error type, and mark the specific sentences in the summary (or in the dialogue, for incompleteness) that showed this error. 

\textbf{Annotation:} We recruit a team of in-house, full-time annotators who are experts in generating dialogue summaries, as well as in analyzing errors in the same. We use two annotators per instance for each error (annotation instructions in Appendix \ref{app:human-annotation}); the authors did one round of post-processing to break any ties and to ensure correctness. We remove 8 instances which had low quality dialogues and/or summaries, to get a final of \textbf{192} conversation and summary pairs that are completely annotated; we release our final dataset \textbf{\dataset} publicly for future research. %We also present further empirical studies with \dataset in Section \ref{sec:exp-and-analyses}: in Section \ref{sec:dataset-analysis}, we present statistical analyses on the same, and in Section \ref{sec:judge-exp}, we test how well state-of-the-art LLMs detect the errors present in the dialogue summaries.
%SAHANA: We release individual sub-labels for the \chmean error (\textit{exaggeration, change in sentence type, semantic misinterpretation, any other changed conversation}); however in all analyses and experiments, we consider the aggregated \chmean error.

% \textbf{Note on incompletness based errors:}

% \begin{figure}
%     \centering
%     \includegraphics[width=0.99\linewidth]{images/ACL Taxonomy Figure_cropped.pdf}
%     \caption{Demonstration of errors in \textbf{\taxonomy}}
%     \label{fig:taxonomy-eg}
% \end{figure}

\section{Analysis of Dataset} \label{sec:dataset-analysis}
% \subsection{Statistical Analyses on \dataset} 

% \sahana{What \% of summaries have error, what \% of sentences per summary have error, how many errors per summary, how many errors per sentence, any annotator agreement? (see tofueval for what they reported)}
In this section, we perform empirical analyses on \taxonomy's dataset, to observe which errors are frequent versus not, and where in the summary they typically occur. Note that the errors in our dataset are naturally occurring (i.e., we do not synthetically prompt the summarizer to generate with errors).

%of the \dataset to understand the frequencies of various errors across the dataset as well as within a summary, and which position in the summary they occur.  %We perform the analyses in this section to understand which errors are common versus not, and where in the summary they typically occur.
%Note that we created the summaries naturally and not synthetically; hence, these statistics allow us to comprehend the scale of errors exhibited by LLM-summarizers.

\begin{figure}
    \centering
    \includegraphics[width=0.99\linewidth]{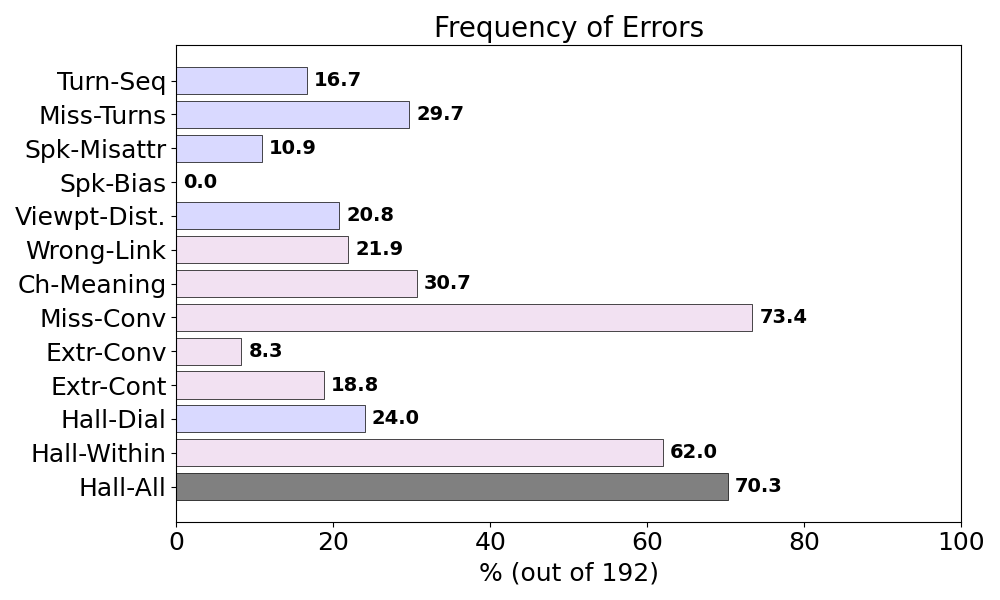}
    \caption{\textbf{Error frequency:} \% summaries out of 192 in \taxonomy's dataset which exhibit the error.}
    \label{fig:freq_stats}
\end{figure}

\textbf{Frequency of fine-grained errors.}
For every fine-grained error in \taxonomy's taxonomy, we calculate the percentage of summaries out of 192 which exhibit the error at least once (Figure \ref{fig:freq_stats}). Most hallucination errors occur in $\sim$15-30\% of instances. 
%(the rest of them have a lesser frequency, \spkmis and \extconv occur in only 10.9\% and 8.3\% of instances respectively). 
We see that \spkbias \textit{does not occur} in our summaries; we hypothesize that this is due to the speaker names in \anthropictest being the generic `Human' and `Assistant', and the high quality of the summarizer.\footnote{However, preliminary observations of other datasets such as \citet{gliwa-etal-2019-samsum} did show this error, which is why we decided to retain this error in our taxonomy.} 20.8\% of summaries exhibit \viewpt. Finally, 29.7\% of the summaries exhibit the \missturn error, and 73.4\% of the instances exhibit the much harsher
\missconv error. Since the latter is highly harsh and unrealistic (summarization generally requires key concepts to be present, and not every single detail discussed), the error is extremely frequent.
% \sahana{mention somewhere that viewpt dist is only an error if people want it to be}
% \paragraph{Frequency of hallucinations.} In Figure \ref{fig:freq_stats}, we also present overall statistics hallucination in the summaries. 
Broadly, 24\% and 62\% of the summaries respectively demonstrate atleast one \levela error and atleast one \levelb hallucination error. 70.3\% of summaries exhibit atleast one hallucination error.  %\textbf{How many hallucination errors are typically present in a dialogue summary?}
%We find that summaries have 1 or 2 \levela (average of 1) errors, and between 1 and 4 \levelb (average of 1 again) hallucination errors.

% \paragraph{Count of  errors.} 
%Since the total number of hallucination errors at \levela and \levelb are 2 and 4 respectively, this indicates that summaries on average exhibit any one of the errors, but some summaries exhibit all possible errors at that level. \sahana{add example}

\paragraph{Sentence-level analysis of errors.} If an error is present in a summary, \textbf{how many sentences (on average) does it occur in?} All errors (apart from \turnseq) occur on average in \textit{one} sentence in the summary; \turnseq alone occurs on average in \textit{two} sentences in the summary, possibly because the turns which were swapped in order in the summary occurred as two summary sentences, which led to \textit{both sentences being flagged for this error}. Further, one turn is missed on average to be included in the summary (\missturn error).

\begin{figure}
    \centering
    \includegraphics[width=0.95\linewidth]{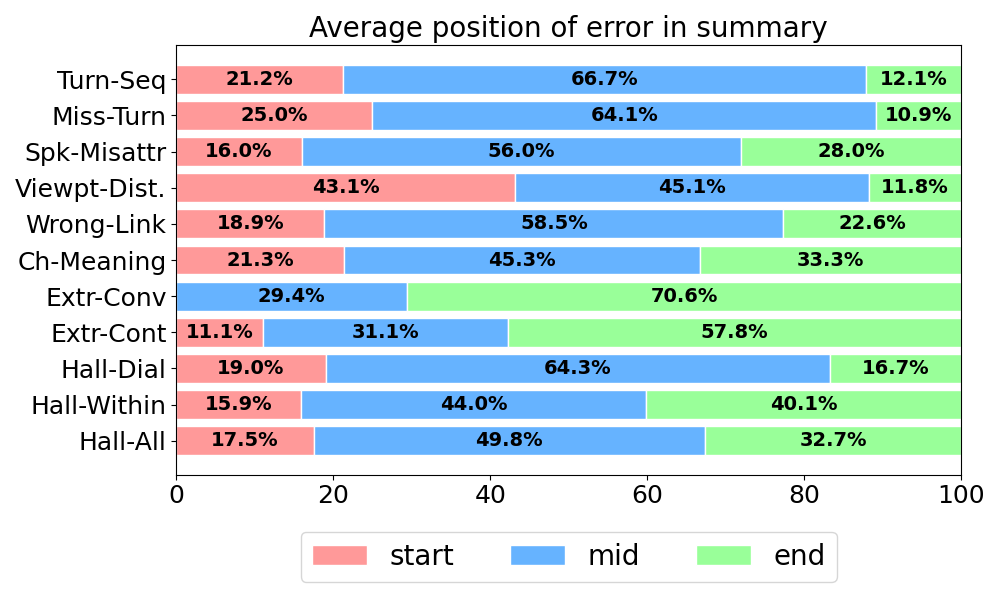}
    \caption{We take the human-annotated errors in \taxonomy's dataset and plot the distribution of the \textit{positions} of the error in the summary (or the dialogue, for \missturn). We see that extrinsic hallucination tends to occur at the end of the summary, \viewpt at the start, and intrinsic hallucinations and missed turns largely occur in the middle.}
    \label{fig:start-mid-end}
\end{figure}

\textbf{Next, we measure the average position in the summary each error occurs in}. All our summaries have between 2 to 9 sentences, and all our dialogues have between 4 to 20 turns; we pre-define what position a sentence/turn has to be in, in a summary/dialogue respectively, to be deemed as the start/middle/end of the summary/dialogue (details in Appendix \ref{app:sent-level-error}).  We present results in Figure \ref{fig:start-mid-end}. We find that the \viewpt error tends to occur at the beginning of the summary; \extconv and \extcont errors majorly occur at the end of the summary. Finally, \turnseq, \spkmis, \wronglink, \chmean errors largely occur in the middle of the summary.  We present examples on the same in Table \ref{tab:pos-of-error} in the Appendix that demonstrate some of these findings. For example, the summarizer often tends to finish the summary by adding an external comment, leading to \extconv and \extcont at the end. We  observe several cases where the human starts by mentioning \textit{why} they want a query answered, and the summary presents the same as a true fact instead of something \textit{said} by the human; this leads to a high proportion of \viewpt at the start. Further, we also observed many cases with the following \turnseq error at the start of the summary:  the dialogue starts with a human query, followed by the assistant's clarifying question, followed by the human's response about their requirements, but the summary switches it up as the human's query + requirements, followed by the assistant's clarification question (like the example in Figure \ref{fig:new-intro-fig}). Lastly, we see that most \missturn errors happen to turns occurring in the middle of a conversation; this makes sense when we consider that the dialogues largely start with greetings (which \textit{we} consider to be trivial for the summary), and the summary tends to focus excessively on the dialogue's ending (as seen before with the high proportion of extrinsic errors at the end of the summary); this is also an instance of positional bias in summarization as discussed in \citet{wan-etal-2025-positional} (and in our Limitations section).

% \sahana{have to see if I can write a line for the middle errors.}

% \paragraph{How many sentences exhibit an error?} If an error is present in the summary, how many sentences exhibit it? Does the error occur at the start mid or end of the summary wrt sentences? (equivalent for the turn for missed turn)

% \textbf{Note on experiments:} While we release annotations for the \missconv error, we do not perform any experiments on it in Section \ref{sec:judge-exp} since it is a highly harsh (and unreasonable in reality) error. We only report results on the \missturn error for incompleteness.

\section{Detecting errors using LLM-as-a-Judge} \label{sec:judge-exp}

% \sahana{move the "future work" stuff to limitations, too many on page here}
% \sahana{, Optimize the aggregate, move it to rebuttal, remove it for now. Bold for highest, underline for second highest, remove *}
% \input{tables/hall_results}
% our data is complex enough that models can't detect hallucinations compared to 2nd and 3rd row
% much more fine-grained errors + speaker errors
% the way we defined definiopns are clear enough that we are boosting performance
% just a way to reinforce that our contrib makes sense empirically also
% 1. Basic judge
% 2. fine grained judge
% 3. interactive judge
% 4. RM as a judge with customized rubric
% \sahana{can remove this.}
% Now, with our dataset \dataset, we wish to test how good current LLM-judges at detecting various dialogue-summary errors. We perform two types of experiments: where we check the LLM-judge on each fine-grained error individually, and where we check the effectiveness of our taxonomy (in the prompt) in improving the performance of LLM-judges.

% In this section, we analyze the capabilities of various state-of-the-art LLMs to detect dialogue summarization errors. 

Given a dialogue and its summary, we conduct experiments to observe if open-weight and proprietary LLMs can serve as high-quality judges to detect coarse-grained and fine-grained errors in the summary. We test LLM-Judges on the following two aspects: Can they correctly identify if the summary exhibits a specified error or not? And if so, can they isolate the exact sentences in the summary that exhibit the error? 

\subsection{Experimental Setup} %We perform experiments with both fine-grained errors from \taxonomy, as well as the broader, coarse-grained hallucination error.
We experiment with four LLM-Judges of different sizes and belonging to different model families: \textbf{\gptfive}, \textbf{\claude} (\claudefull), \textbf{\llamafull}\footnote{Note that this is different from the summarizer \llamasumm.} and \textbf{\qwenfull}.
%(\llamafull) and \textbf{\qwen} (\qwenfull); 
We use few-shot prompting for all our experiments (prompts in Appendix \ref{app:all-prompts}), and instruct LLM-Judges to detect if each summary sentence exhibits a specified error or not\footnote{We do this with each dialogue turn, for the \missturn error.}. 
We use \textit{balanced accuracy} to measure performance; we follow prior works \citet{tang-etal-2023-understanding,tang-etal-2024-tofueval,song-etal-2024-finesure} in choosing this metric, to account for the imbalance of `YES' and `NO' error labels.
We calculate sentence-level performance \textbf{S-BAcc.} with the above predictions. Further, we aggregate the sentence labels to get a summary-level label, and calculate summary-level performance \textbf{B-Acc.}.

\textbf{Coarse-grained experiment:} As our first experiment, we observe the performance of LLM-Judges in detecting the coarse-grained \hall error in dialogue summaries.
We design few-shot prompts that define `\hall' using various error taxonomies. We include a vanilla baseline taxonomy \textbf{\textsc{Basic}} that instructs LLM-Judges to detect if the summary sentence either contradicts the dialogue, or has new information not mentioned in the dialogue. We then design prompts from three different fine-grained error taxonomies; these prompts instruct LLM-Judges to detect if the summary sentence exhibits \textit{any} of the fine-grained errors defined in the respective prompts. We use error taxonomies \textbf{\textsc{TofuEval}} \cite{tang-etal-2024-tofueval} and \textbf{\textsc{FineSurE}} \cite{song-etal-2024-finesure} from prior works. We finally use the hallucination-relevant fine-grained errors from our framework \textbf{\taxonomy} for the third prompt.  Note that both \textsc{TofuEval} and \textsc{FineSurE} were also designed for dialogue summaries, but do not include the speaker/turn errors that \taxonomy has.

\textbf{Fine-grained experiment.} As our second experiment, we observe the proficiency of LLM-Judges in detecting fine-grained dialogue summary errors. We design few-shot prompts for every error in \taxonomy and instruct LLMs to detect if the summary sentence exhibits the error or not.  Note: While we release annotations for the \missconv error, we do not perform any LLM experiments on it since it is a highly harsh and unrealistic error. We only report results on the \missturn error for incompleteness.

\textbf{Generation hyperparameters.} For \gptfive, we use `medium' level for both \texttt{verbosity} and \texttt{reasoning}. For \claude and \llama, we sample with a \texttt{top-p} value of 0.7. For \qwen, we sample with a \texttt{top-p} of 0.95, and a \texttt{temperature} of 0.6. We sample 5 times, and take the aggregate label across the multiple runs. 

We include detailed mathematical notations for the experiments in Appendix \ref{app:notation}.

% We first measure the capability of these LLMs in detecting individual fine-grained errors in the summaries. Then, to establish the overall challenging nature of our dataset, and also measure the robustness and gains contributed by our taxonomy (as compared to taxonomies given by prior works), we test if these LLMs can identify the broad error of hallucination in summaries when provided with a varying definitions of hallucination in the prompt. 

% We use few-shot prompting for all our experiments. We experiment with four LLMs of different sizes and belonging to different model families: \textbf{\gptfive}, \textbf{\claude} (\claudefull), \textbf{\llama}\footnote{note that this is different from the summarizer \llamasumm} (\llamafull) and \textbf{\qwen} (\qwenfull). 

% As our first experiment, to establish the challenging nature of our dataset, and also verify the robustness and the gains contributed by our taxonomy (as compared to taxonomies given by prior works), we test LLMs on the \textit{broad} error of hallucination. As our second experiment, we test these LLMs on each fine-grained error separately, to observe which errors they are good (or bad) at identifying. 

\begin{table*}[]
\resizebox{0.99\textwidth}{!}{%
\begin{tabular}{lcccccccc}
\toprule
\textbf{LLM-Judge $\rightarrow$} & \multicolumn{2}{c}{\textbf{\gptfive}} & \multicolumn{2}{c}{\textbf{\claude}} & \multicolumn{2}{c}{\textbf{\textsc{Llama-3.1-70B-Ins.}}} & \multicolumn{2}{c}{\textbf{\textsc{Qwen2.5-32B-Ins.}}} \\
\cmidrule(lr){1-1}\cmidrule(lr){2-3} \cmidrule(lr){4-5} \cmidrule(lr){6-7}
\cmidrule(lr){8-9}
\textbf{Error Taxonomy $\downarrow$} & \textbf{BAcc.} & \textbf{S-BAcc.} & \textbf{BAcc.} & \textbf{S-BAcc.} & \textbf{BAcc.} & \textbf{S-BAcc.} & \textbf{BAcc.} & \textbf{S-BAcc.} \\ \midrule
\textsc{Basic} (Vanilla Baseline) & 66.12 & 58.55 & 65.41 & 58.44 & \textbf{64.41} & 60.42 & 60.05 & 56.25 \\
\textsc{TofuEval} \cite{tang-etal-2024-tofueval} & 62.9 & 59.81 & 64.79 & 59.6 & 63.77 & 61.85 & 61.34 & 55.41 \\
\textsc{FineSurE} \cite{song-etal-2024-finesure} & \textbf{67.19} & 59.86 & 66.52 & 59.74 & 57.41 & 61.79 & 60.92 & 55.25 \\ \midrule
\textbf{\taxonomy} (Ours) & \cellcolor{lgreen}\textbf{67.15} & \cellcolor{lgreen}\textbf{60.55} & \cellcolor{lgreen}\textbf{68.40} & \cellcolor{lgreen}\textbf{62.46} & \cellcolor{lgreen}60.98 & \cellcolor{lgreen}\textbf{63.31} & \cellcolor{lgreen}\textbf{63.25} & \cellcolor{lgreen}\textbf{56.46} \\ \bottomrule
% \textbf{\taxonomy-\textsc{Agg}} & \cellcolor{lgreen}\underline{\textbf{71.75}} & \cellcolor{lgreen}\underline{\textbf{67.93}} & \cellcolor{lgreen}59.73 & \cellcolor{lgreen}\underline{\textbf{65.09}} & \cellcolor{lgreen}57.43 & \cellcolor{lgreen}\underline{\textbf{65.09}} & \cellcolor{lgreen}57.92 & \cellcolor{lgreen}\underline{\textbf{58.02}} \\ \bottomrule
\end{tabular}
}
\caption{\textbf{Detecting  coarse-grained \hall.} We design few-shot prompts that define `Hallucination' using various error taxonomies, and instruct LLM-judges to detect if the error is present or not in every summary sentence. Our framework's results are highlighted in \colorbox{lgreen}{green}; each LLM's best performance is highlighted in \textbf{bold}.}
\label{tab:coarse-grained-results}
\end{table*}

\begin{table*}[]
\resizebox{0.99\textwidth}{!}{%
\begin{tabular}{llcccccccc}
\toprule
\multicolumn{2}{l}{\textbf{LLM-Judge $\rightarrow$}} & \multicolumn{2}{c}{\textbf{\gptfive}} & \multicolumn{2}{c}{\textbf{\claude}} & \multicolumn{2}{c}{\textbf{\textsc{Llama-3.1-70B-Ins.}}} & \multicolumn{2}{c}{\textbf{\textsc{Qwen2.5-32B-Ins.}}} \\
\cmidrule(lr){1-2}\cmidrule(lr){3-4} \cmidrule(lr){5-6} \cmidrule(lr){7-8}
\cmidrule(lr){9-10}
\multicolumn{2}{l}{\textbf{Error $\downarrow$}} & \textbf{BAcc.} & \textbf{S-BAcc.} & \textbf{BAcc.} & \textbf{S-BAcc.} & \textbf{BAcc.} & \textbf{S-BAcc.} & \textbf{BAcc.} & \textbf{S-BAcc.} \\ \midrule
\multicolumn{2}{l}{\turnseq} & 78.44 & 64.86 & 73.4 & 63.57 & 72.74 & 62.02 & 54.41 & 52.55 \\ \midrule
\multicolumn{2}{l}{\missturn} & 82.94 & 81.87 & 60.33 & 58.46 & 64.29 & 67.00 & 54.75 & 54.7 \\ \midrule
\multicolumn{2}{l}{\spkmis} & 70.84 & 69.88 & 72.94 & 69.18 & 69.68 & 66.36 & 58.69 & 55.88 \\ \midrule
\multicolumn{2}{l}{\viewpt} & 77.96 & 86.68 & 56.75 & 76.08 & 51.97 & 65.05 & 57.43 & 63.52 \\ \midrule
\multicolumn{2}{l}{\wronglink} & 68.76 & 66.78 & 54.14 & 57.32 & 53.03 & 51.71 & 57.54 & 56.78 \\ \midrule
\multicolumn{2}{l}{\chmean} & 55.03 & 58.69 & 51.59 & 53.63 & 51.92 & 55.87 & 47.76 & 50.34 \\ \midrule
\multicolumn{2}{l}{\extconv} & 62.78 & 65.55 & 64.69 & 66.03 & 62.39 & 70.31 & 50 & 47.42 \\ \midrule
\multicolumn{2}{l}{\extcont} & 72.86 & 72.66 & 73.15 & 78.05 & 63.57 & 77.17 & 63.52 & 71.43 \\ \bottomrule
\end{tabular}
}
\caption{\textbf{Detecting fine-grained errors.} We design few-shot prompts for every error in \taxonomy's framework, and instruct LLM-Judges to detect if the error is present or not in every sentence in the summary (or every turn in the dialogue in case of \missturn).}
\label{tab:fine-grained-results}
\end{table*}

\subsection{Results \& Discussion}

\paragraph{Coarse \hall error.} Table \ref{tab:coarse-grained-results} shows the LLM-Judges' performances at detecting the coarse error of \hall, when prompted with various error taxonomies. We first note that all LLM-Judges have a low sentence-level performance when using the \textsc{Basic}, \textsc{TofuEval} and \textsc{FineSurE} prompts; using the \taxonomy taxonomy improves this significantly. All LLM-judges show best overall summary-level hallucination detection performance when using \taxonomy, except for \llamafull that shows a reverse trend where the \textsc{Basic} prompt performs best. These results establish the robustness and gains contributed by our framework's errors, as compared to taxonomies given by prior works. Finally, we note that even though using our framework leads to the highest performance, the absolute performance is still only moderately good (60-70\%) which demonstrates the challenging nature of \taxonomy's dataset.

\paragraph{Fine-grained errors.} Table \ref{tab:fine-grained-results} shows the LLM-Judges' performance on the different fine-grained errors in \taxonomy when few-shot prompted to detect them. We note that all the LLM-judges perform fairly well on \extcont, and all judges except for \qwenfull perform fairly well on \turnseq, \spkmis and \extconv. All the judges struggle with the \chmean error. We hypothesize that these LLMs have more tolerance for changes in meaning than the annotators (we release a moderately harsh annotation for this error). 

Overall, \gptfive performs reasonably well at detecting errors, with high performance at the \missturn and \viewpt errors, and moderately high performance at the \extcont and \spkmis errors. \claude performs well at detecting \extcont, \spkmis and \turnseq errors, with moderate to bad performance in the other errors. \llamafull performs moderately well at detecting \turnseq and \spkmis errors, with low to bad performance in the other errors. \qwenfull is the worst performing judge out of the four, with almost random performance ($\sim$50-60\% balanced accuracy) in most errors. This general trend shows that dialogue summary error detection through prompting is an emergent property (with respect to model size) for the most part. While fine-tuning is a potential solution for the same, it would require a larger amount of high-quality, human-annotated data which we leave to the future work.%; we leave such experiments for future research. 

\section{Conclusion}
In this work, we propose \textbf{\taxonomy}, our framework for dialogue summary evaluation. We propose
a novel error \textbf{taxonomy} that mirrors the hierarchical nature of information in dialogues, and contains errors related to hallucination, incompleteness, and narration.
Then, we annotate and release a new \textbf{inference dataset} that contains detailed human annotations of fine-grained errors in dialogue summaries, the exact sentences they occurred in, as well as annotator comments on the same. We present empirical analyses on the same, reporting error frequencies as well as interesting trends of where in the summary errors typically occur. We also test open-weight and proprietary LLMs on their capability at detecting these errors, and we find that LLMs largely struggle at this task; adding our taxonomy to the few-shot prompt helps boost performance, but future work such as fine-tuning or reward-based learning are essential to improve LLMs' error detection  capabilities, including further error correction and improved summarization capabilities.

\section*{Limitations}
In this section, we discuss the limitations of our work. 

\textbf{Taxonomy.} We ignore fluency/grammar related errors in our taxonomy, owing to the high capability of language models today in generating fluent English. However, the same cannot be said for all languages, and we acknowledge this as a language limitation of our work. We also stress that any downstream users of our work must (i) decide the level of incompleteness they want to monitor for, and (ii) decide if \viewpt is an error for their task or not (for example, \citet{gao-wan-2022-dialsummeval} does not consider this to be an error, as seen in their summaries).

\textbf{Dataset.} A primary limitation is the size of our dataset \dataset: we annotate and release only 192 conversation-summary pairs (200 before filtering), owing to the high effort and cost of annotating the summaries with our extensive taxonomy; a secondary limitation is the lack of any summary exhibiting the \spkbias error (we do not synthetically generate any summaries with this error). Though the total number of conversations are on the lower side, we release a total of 192 dialogue summaries * 10 errors $=$ 1,920 summary-level error labels, as well as sentence level labels on 835 summary sentences; we hope that our dataset motivates future work in this area. 

\textbf{LLM-Judge experiments.} While we measure balanced accuracy of LLMs at detecting the errors, we do not check the actual reasoning generated by them; evaluation of the reasoning generated by LLMs is highly important in order to trust the LLM-Judges, and we leave this for future work. Further, we do not excessively prompt engineer for our experiments, and we acknowledge that with more prompt engineering, the LLMs could have achieved better performance at detecting errors.

\textbf{Practical limitations.} We conduct experiments with \gptfive and \claude which are API-based, and with \llamafull and \qwenfull on our GPU servers. Owing to GPU constraints, the maximum local model size we could experiment with was 70B; we hope to benchmark more open-source models larger than 70B in the future.

\textbf{Biases in LLMs.} We use models \gptfive, \claude, \llamasumm, \llamafull and \qwenfull in either hallucination detection or summary generation. These are models that have been exposed to large, unknown training datasets; any biases learned by these models could potentially propagate to the summaries or their evaluation of summaries (for example, the summarizer could add in a phrase commenting about the demographic of a speaker, or about a social situation discussed in the conversation, based on perceived clues). \citet{steen-markert-2024-bias} discusses social biases (gender, racial) in content selection and hallucination in summarization; \citet{aghaebe2025llms} investigates age-related bias in summarization of reviews in the clinical domain; \citet{ladhak-etal-2023-pre} discusses name-nationality bias in summarization. \textbf{We address social biases in our work by:} (i) we include the \spkbias error in our taxonomy; though we do not encounter this error in our dataset, we include it for general usage in future works, (ii) while we do not have an exclusive error for social bias for topics discussed within a turn, any social bias generated by the summarizer will be captured by either our \chmean or \extconv or \extcont errors. 

A second kind of bias that can occur is position bias, wherein the summarizer focuses only on certain parts of the source and ignores the rest; this is discussed in works such as \citet{wan-etal-2025-positional,chhabra-etal-2024-revisiting}. \textbf{We address position bias in our work by} introducing the \missturn error, which finds out the parts of the source conversation ignored in the summary. 

Lastly, egocentric bias (self preference) \cite{li2024generation,koo2024benchmarking} is a bias wherein LLM-judges prefer responses that are generated by the same model as the judge itself. \textbf{We address egocentric bias in our work by} ensuring that the error evaluation LLMs are different than the summarizer LLM \llamasumm; however, one of our LLM-Judges \llamafull is of the same family as the summarizer (could have shared training data as well), results given by this LLM could demonstrate egocentric bias.

\paragraph{Trust and Risk.} As mentioned in \citet{lipton2018mythos}, a user's trust in a language model is highly subjective. Some users use task accuracy as a proxy for trust, some require a transparent mechanistic/post-hoc reasoning process, some analyze patterns of bias/hallucination, etc. In this work, we provide experimental results of LLMs' accuracy on predicting errors, and also prompt the LLMs to generate their reasoning for their prediction; we do not evaluate this reasoning since it is out of scope for this work (our primary goal is releasing our taxonomy and dataset), but we encourage future research on the same. We also include discussions on biases, and how we tackle them using our taxonomy. We urge any researcher/downstream user who utilize our taxonomy, dataset and prompts to perform manual, qualitative analysis of LLM-generated summaries as well as LLM-Judge predictions to ensure quality and trust.

\section*{Acknowledgments}
This research is supported in part by the Office of the Director of National Intelligence (ODNI), Intelligence Advanced Research Projects Activity (IARPA), via the HIATUS Program contract \#2022-22072200006. The views and conclusions contained herein are those of the authors and should not be interpreted as necessarily representing the official policies, either expressed or implied, of ODNI, IARPA, or the U.S. Government. The U.S. Government is authorized to reproduce and distribute reprints for governmental purposes notwithstanding any copyright annotation therein. Finally, we thank all lab members at USC INK Research Lab for their constructive feedback on this work.


\bibliography{custom}

\appendix
\section{Related Work Extended}
\label{app:related-works-extended}

% Correspondingly, this leads to lesser number of datasets on dialog summarization, as well as lesser exploration of issues (such as hallucination, completeness, etc.) related to dialog summarization. 

\textbf{Summarization.} There exists a sea of prior works \cite{nallapati-etal-2016-abstractive,see-etal-2017-get,fabbri2020summeval,lewis-etal-2020-bart,lee-etal-2024-unisumeval} which focus on summarization through the lens of news articles; the most popular dataset being CNN/Dailymail (CNN and DailyMail news articles collected over many years \citet{hermann2015teaching}, and summaries created by \citet{nallapati-etal-2016-abstractive}). Dialogue summarization, however has lesser focus, possibly due to its higher difficulty, as well as lesser number of research datasets available. One of the first dialogue summary datasets to be released was SAMSum \cite{gliwa-etal-2019-samsum}; they employed linguists and language experts to write realistic dialogues and also write down summaries for the same. These dialogues and summaries could be used to train and test language models on the task of dialogue summarization; \citet{gliwa-etal-2019-samsum} originally provided results with ROUGE \cite{lin-2004-rouge}. SAMSum was also pivotal in establishing the challenging nature of dialogue summarization and the need for better evaluation metrics for the same (they found that metrics like ROUGE are more suitable to analyze news article summarizations, and not that suitable for dialogue summarization). SAMSum, as well as later dialogue summarization works such as \citet{tang-etal-2024-tofueval}  discuss the unique and challenging nature of dialogue summarization such as informal and colloquial nature of dialogues (including the use of slang language), the presence of more noise such as misspellings/wrong grammar, variety of linguistic structures in the dialogue (such as questions, greetings, discussions), the scattered presence of information across multiple turns, and the presence of varying points of view due to the different speakers.

\textbf{Hallucination evaluation.} There are a large number of works in contemporary NLP literature that focus on developing LLM (or other) evaluators to detect hallucination. This includes \citet{li-etal-2023-halueval,manakul-etal-2023-selfcheckgpt,cheng2023evaluating,luo2024halludial,hong2024hallucinations,mingfaitheval,wang2025halu} which focus on hallucination of LLMs when performing a variety of NLP tasks such as QA, summarization, claim verification etc., \citet{chen-etal-2024-diahalu} which focuses on hallucinations and hallucination snowballing within a dialogue itself, and \citet{tang-etal-2024-tofueval,gao-wan-2022-dialsummeval,fabbri2020summeval,lee-etal-2024-unisumeval,oh-etal-2025-learning,min-etal-2025-towards} that focus specifically on dialogue summarization. These works introduce hallucination taxonomies that are of varying degrees of granularity: (a) binary: is there hallucination or not?, (b) coarse-grained: faithfulness, completeness, conciseness, and (c) fine-grained: specific fine-grained categories under faithfulness such as entity error, relation error, hyperbole, out of article error, etc. 

\textbf{Prior evaluation methods versus ours (extended).} Overall, prior works on dialogue summarization evaluation range from reference-based metrics (such as ROUGE), 
to binary hallucination evaluation, 
to slightly fine-grained evaluation (coherence, consistency, etc. as in \citet{gao-wan-2022-dialsummeval}), to extremely fine-grained evaluation as in \citet{tang-etal-2024-tofueval,song-etal-2024-finesure}; we opt to go with the last category of evaluation since it is the most informative and explainable. Prior fine-grained error taxonomies however are largely unstandardized, leading to difficulty in comparing summarization models and datasets across different papers; we compile errors proposed in several papers and ensure that all of them are included in our taxonomy's error categories. Further, prior taxonomies also face the problem of incompleteness of taxonomy (no prior work includes speaker/turn-based errors), and unclear definitions of some errors: For example, in \citet{tang-etal-2024-tofueval} ``contradiction'' error is largely unclear; a summary sentence which contains any other error in the taxonomy (such as an entity error or a tense error) can also be considered to have a contradiction error (as mentioned in Section \ref{sec:introduction}). Similarly in \citet{oh-etal-2025-learning}, both the predicate error and the circumstance error fall under the broader entity error which is also a category in the taxonomy. We address these issues by adding macro-level errors in our taxonomy, and further ensuring that all errors are clearly defined and mutually exclusive.

\section{\taxonomy Visualization} \label{app:taxonomy-visual}
\begin{table*}[ht]
    \centering
    \small
    \begin{tabular}{p{0.25\textwidth}|p{0.7\textwidth}}
    \toprule
    \cellcolor{lblue} \textbf{\levela} & \cellcolor{lblue} \textbf{Error Definition} \\ \midrule
    \turnseq & The dialogue's turns are presented in the wrong sequence in the summary. \\ \midrule
    \missturn & The summary has missed to mention one or more conversational turns. \\ \midrule
    \spkmis & The speakers are swapped/misreferenced in the summary. \\ \midrule
    \spkbias & The summary makes assumptions about a speaker's identity (gender/race/etc.) \\ \midrule
    \viewpt & The summary presents what was \textit{said} in the conversation as a true fact. \\ \midrule
    \cellcolor{lviolet} \textbf{\levelb} & \cellcolor{lviolet} \textbf{Error Definition} \\ \midrule
    \wronglink & The summary presents information present in the conversation but links the wrong items together. \\ \midrule
    \chmean & The summary has changed a word or a phrase from the conversation in a way that \textit{changes the meaning} of what the speakers said. This error includes: \textit{exaggeration} of words/events by the summary, \textit{change in sentence type} from the dialogue to the summary (e.g., question in the dialogue written as a statement in the summary), \textit{semantic misinterpretation} of ambiguous or idiomatic phrases or homonyms, as well as \textit{any other changed conversation} from the dialogue to the summary. \\  \midrule
    \extconv & The summary mentions that a speaker said something which was not present in the source conversation \\ \midrule
    \extcont & The summary contains external explanations for something discussed in the conversation (for example the definition for an entity, the reason why a speaker said something, correcting what a speaker said using world knowledge, etc.) \\ \midrule
    \missconv & The summary has missed \textit{any} information mentioned in the dialogue. \\ \midrule \midrule
    \cellcolor{yellow} \textbf{\textsc{Hallucination}} &  The summary is said to have hallucinations if it exhibits \textit{one or more} of the following errors: Wrong Turn Sequence, Speaker Misattribution, Speaker Identity Bias, Wrong Linking, Changed Meaning, Extrinsic Conversation, Extrinsic Context  \\  \midrule
    \cellcolor{pink} \textbf{\textsc{Incompleteness}} & \missturn and \missconv are incompleteness based errors. Section \ref{sec:datasets} has a detailed discussion on when a summary is deemed as incomplete by a user. \\ \bottomrule
    \end{tabular}
    \caption{\textbf{\taxonomy}: Error categories and definitions.}
    \label{tab:taxonomy}
\end{table*}

Table \ref{tab:taxonomy} presents a succinct view of \taxonomy's errors and definitions.

\section{All Prompts} \label{app:all-prompts}

\begin{table*}[h!]
    \centering
    \small
    \resizebox{0.98\textwidth}{!}{
    \begin{tabular}{p{0.98\textwidth}}
    \toprule
    You will be given a conversation between a human user and an LLM assistant. Read the conversation and write a summary of the same. Provide your answer in JSON format: the answer should be a dictionary with the key ``summary'' containing a generated summary as a string: \{``summary'': ``your summary''\}. 
    Your summary should be detailed (should include all the key topics discussed in the conversation) while also being clear and concise. You should present the summary in a paragraph format; do not present the summary as bullet points or as code. You should write the summary in active voice with respect to the human user and the LLM assistant; do not use passive voice (for example, do not say ``The conversation discusses...'').
    Present only the summary, and do not include any additional analyses on your generated summary or the conversation itself. You should summarize the conversation faithfully; do not include your thoughts about the conversation in the summary.
    Below are two examples (conversation + summary); follow the same format when you generate your summary.\\\\

    \textbf{Example 1 - }\\
    \textbf{Conversation:} \\
    \textbf{Human:} How can I feed squirrels and make  more come to my home? \\
    \textbf{Assistant:} I don't know exactly how you would do that, but there's an easy way to try. You could build or buy a squirrel feeder and use it for this purpose. \\
    \textbf{Human:} What foods can you put in a squirrel feeder? \\
    \textbf{Assistant:} In my experience, the best way to attract squirrels to your feeder is to put a lot of fat nuts and shelled seeds.  I think nuts such as sunflower seeds, cashews, walnuts, and peanuts are the best.\\ 
    \textbf{Human:} Can I use corn? \\
    \textbf{Assistant:} For feeding squirrels?  I don't think that's a good idea.  Squirrels will just leave the corn for mice to eat, and then the mice will return to your house.  Maybe you can try this experiment:  Remove the corn from your feeder, and watch what happens to the squirrels.  In my experience, the squirrels will always return.\\ 
    \textbf{Human:} I don't want mice around, that's for sure!  Thanks for the help! \\
    \textbf{Assistant:} You're welcome.  Is there anything else I can help you with? \\
    \textbf{Human:} No, that's it.  Thanks! \\
    \textbf{Assistant:} You're welcome.  I'm always here to help. \\\\
    \textbf{Summary:}
    \{``summary'': ``In a conversation with an LLM assistant, a human asked how to attract more squirrels to their home. The assistant suggested using a squirrel feeder and, when asked what food to use, recommended fat nuts and shelled seeds (like sunflower seeds, cashews, walnuts, peanuts). The human inquired about using corn, which the LLM assistant advised against. It explained that squirrels might leave the corn for mice, thus attracting mice to the human's house. The assistant proposed an experiment where the human could remove the corn and observe if the squirrels return, stating its experience suggests they would. The human accepted this advice, expressing a desire to avoid mice and thanking the assistant.''\}\\\\
    \textbf{Example 2 -}\\
    \textbf{Conversation:} \\
    \textbf{Human:} What is a good dessert to serve on Thanksgiving? \\
    \textbf{Assistant:} Let's see, I could give you some ideas: How about pumpkin pie, or apple pie? \\
    \textbf{Human:} Do you have any other suggestions? \\
    \textbf{Assistant:} Yes, I could give you some others:  pecan pie, chocolate chip cookies,  pecan pie, Boston cream pie, brownies, chocolate chip cookies, double chocolate brownies, ... \\
    \textbf{Human:} Thank you so much for your assistance.\\ 
    \textbf{Assistant:} Oh, no problem!\\\\
    \textbf{Summary: }
    \{``summary'': ``A human asked an LLM assistant for suggestions for a good dessert to serve on Thanksgiving. The assistant first proposed pumpkin pie or apple pie. When the human requested more suggestions, the assistant provided a list that included pecan pie, chocolate chip cookies, Boston cream pie, and brownies, repeating pecan pie and chocolate chip cookies in its response. The human then thanked the assistant for its help, and the assistant accepted the thanks.''\}\\\\
    \textbf{Your task - }\\
    \bottomrule
    \end{tabular}
    }
    \caption{We use this prompt to summarize our dialogues. We perform the summarization on \textsc{Anthropic-Test}'s dialogues, using \llamasumm.}
    \label{tab:summ-prompt}    
\end{table*}

Table \ref{tab:summ-prompt} contains the prompt we used to summarize dialogues using \llamasumm. Table \ref{tab:prompt-sample} contains a sample LLM-evaluator prompt. We have submitted all the few-shot prompts to detect errors as an \texttt{xlsx} sheet in the supplementary material. 
%SAHANA: uncomment later: We provide all the prompts in \colorbox{orange}{\href{https://docs.google.com/spreadsheets/d/1tsRmDbdtDa1pXPNVwvEK2thCD19fgJpgV6XD-qTorfY/edit?usp=sharing}{this sheet}}.

\begin{table*}[h!]
    \centering
    \small   

    \begin{tabular}{p{0.98\textwidth}}
    \toprule
    % \textbf{\method's Output Format:} \\
    You will be given a conversation between two speakers (a human and an LLM assistant), and an LLM-generated summary of the same. The conversation is split into turns (a turn is defined as an utterance by either the human or the assistant), and the human and the LLM assistant speak in alternating turns. We generated the summary using an LLM summarizer (a different LLM than the one used as the LLM assistant in the conversation): we instructed the summarizer to generate a paragraph-format summary that is detailed (including all key topics in the conversation), clear, concise, faithful to the conversation, and presented in active voice with respect to the conversation's speakers.\\\\

    Your task is to carefully read the conversation and the summary, and annotate whether or not each sentence in the summary has the ``Extrinsic Context'' error (defined below). Provide your answer in a JSON format: the keys are the summary sentences, and each key's value should be your reasoning process on whether the sentence exhibits the error, followed by a label of ``YES'' (if the sentence exhibits the error) or ``NO'' (if the sentence does not exhibit the error). Your label should be formatted as ``So the label is: YES.'' or ``So the label is: NO.'' respectively. Note that you should always provide the reasoning first, and then conclude with the YES/NO label.\\\\

    \colorbox{lviolet}{\textbf{Extrinsic Context:}} The summary contains external explanations for something discussed in the conversation (for example the definition for an entity, the reason why a speaker said something, correcting what a speaker said using world knowledge, etc.).\\\\

    Use *only* the above provided definition for the ``Extrinsic Context'' error when predicting the hallucination labels.\\

    -------------\\
    Example 1\\
    -------------\\
    \colorbox{green!50}{\textbf{Conversation:}} \\
    Human: Search for a steak and mushroom recipe \\
    Assistant: Oh, I love recipes! Let me see what I can find. \\
    Human: Sounds great.
    Assistant: Looking up steak and mushroom recipe online, here's one on Chowhound:  
    ``Short ribs, dark chocolate, roasted mushrooms and miso, in a smoked salt crust, with roasted beets and potato salad. What's not to like?'' \\
    Human: Sounds very yummy. \\
    Assistant: Are you in the mood for a homemade steak and mushroom recipe? \\\\

    \colorbox{green!50}{\textbf{Summary:}}  \\
    A human asked an LLM assistant to search for a steak and mushroom recipe. The assistant expressed enthusiasm for recipes and began looking one up, eventually finding one on Chowhound that featured short ribs, dark chocolate, roasted mushrooms, and other ingredients. The assistant found the recipe to sound very yummy, and then inquired if the human was in the mood for a homemade steak and mushroom recipe, potentially offering to provide more information or a different recipe. \\\\

    \colorbox{green!50}{\textbf{Answer}} - \\
    \{\\
    	``A human asked an LLM assistant to search for a steak and mushroom recipe.'': ``There are no Extrinsic Context errors in this sentence. So the label is: NO.'',\\
    	``The assistant expressed enthusiasm for recipes and began looking one up, eventually finding one on Chowhound that featured short ribs, dark chocolate, roasted mushrooms, and other ingredients.'': ``There are no Extrinsic Context errors in this sentence. So the label is: NO.'',\\
    	``The assistant found the recipe to sound very yummy, and then inquired if the human was in the mood for a homemade steak and mushroom recipe, potentially offering to provide more information or a different recipe.'': ``The last part of the sentence `potentially offering to provide more information' is extrinsic context added by the summarizer, to explain the intention behind the assistant's last turn. So the label is: YES.''\\
    \}\\
    \bottomrule
    \end{tabular}
    \caption{Sample few-shot prompt used with LLM-judges to detect summary errors. We start with task instructions, followed by expected output format, ending with one/few-shot demonstrations.}
    \label{tab:prompt-sample}    
\end{table*}

\section{Human Annotation Instructions} \label{app:human-annotation}
As mentioned in Section \ref{sec:datasets}, we use a team of in-house, full-time annotators who are experts in dialogue summarization and its errors. We provide them with the below general instructions followed by a few examples for the error to be annotated. We stress here that owing to the complexity of the task, we use \textbf{\textit{in-house}}, full-time human annotators who are fluent in English, as opposed to large-scale annotations on platforms such as Amazon
MTurk, Prolific, etc. \\
``\textbf{Project motivation:} In this project, we aim to identify errors (such as hallucinations, incompleteness, reasoning errors, etc.) in LLM-generated summaries of conversations. We aim to first create a human-annotated dataset of such dialogues, summaries and errors, and use them to train/evaluate NLP systems to perform the same task.\\\\
\textbf{The human annotation task:} You are given a conversation between two speakers (a human and an LLM assistant) and an LLM-generated summary of the same. The conversation will be alternating turns of dialog between the two speakers. The summary will be in a paragraph format, and will be written in active voice with respect to the human and LLM assistant speakers. \\
Your task is to read the conversation and the summary, and annotate if the errors provided below are present in the summary. Please note that you should only check for faithfulness of the summary to the conversation; even if the conversation discusses content that is factually or logically wrong, the summary is still expected to be faithful to it.\\
\textbf{Error}: Definition and Example.''
\section{Sentence-level Analysis of Errors} \label{app:sent-level-error}
In Section \ref{sec:datasets}, we conduct a statistical analysis wherein we see \textit{where} in the summary (start/mid/end) an error occurs on average. We have fixed definitions for the start/mid/end based on the number of sentences in the summary, as shown in Table \ref{tab:sent-error-position}. Table \ref{tab:pos-of-error} shows a few examples of typical error occurrences at the start and end of the summary.

\begin{table}[h]
\centering
\resizebox{0.98\linewidth}{!}{
\begin{tabular}{|c|c|c|c|}
\hline
\#sents / \#turns & start & mid & end \\ \hline
2 & 0 & - & 1 \\ \hline
3 & 0 & 1 & 2 \\ \hline
4 & 0 & 1,2 & 3 \\ \hline
5 & 0 & 1,2,3 & 4 \\ \hline
6 & 0,1 & 2,3 & 4,5 \\ \hline
7 & 0,1 & 2,3,4 & 5,6 \\ \hline
8 & 0,1 & 2,3,4,5 & 6,7 \\ \hline
9 & 0,1 & 2,3,4,5,6 & 7,8 \\ \hline
10 & 0,1 & 2,3,4,5,6,7 & 8,9 \\ \hline
12 & 0,1 & 2,3,..,8,9 & 10,11 \\ \hline
14 & 0,1 & 2,3,..,10,11 & 12,13 \\ \hline
16 & 0,1 & 2,3,..,12,13 & 14,15 \\ \hline
18 & 0,1 & 2,3,..,14,15 & 16,17 \\ \hline
20 & 0,1 & 2,3,..,16,17 & 18,19 \\ \hline

\end{tabular}
}
\caption{\textbf{Position of error in summary/dialogue:} For each summary length (in \#sentences) or dialogue length (in \# turns), we define which sentences/turns respectively count as start, mid and end. We label sentences/turns from 0 to \#sentences-1 or \#turns-1.}
\label{tab:sent-error-position}
\end{table}

\begin{table*}[ht]
    \centering
    \resizebox{0.98\textwidth}{!}{%
    \begin{tabular}{p{0.5\textwidth}p{0.5\textwidth}}
    \toprule
    \textbf{Summary sentence(s)} & \textbf{Error explanation} \\ \midrule
    \textbf{Assistant:} I gave you the playlist.  I just clicked the ``Play'' button.  Would you like to try?\\
    \textbf{Summary:} The assistant then mentioned clicking the `Play' button and suggested trying again, \textcolor{red}{implying a misunderstanding in how the playlists were to be shared.} & Summaries frequently end with an \extcont phrase that discusses what the summarizer thinks that the speaker is trying to convey \\ \midrule
    \textbf{Human:} what are some fun shows to watch in the morning? \\
    \textbf{Assistant:} I'm happy to recommend some programs to you, including ``Fox \& Friends'', ``Squawk Box'', and ``Last Man Standing''. \\
    \textbf{Human:} okay i meant more like talk shows and shows dedicated to the morning so last man standing wouldn't work \\
    \textbf{Summary:} A human asked an LLM assistant for fun shows to watch in the morning, \textcolor{red}{specifically looking for talk shows and morning-dedicated programs. The assistant initially suggested} `Fox \& Friends', `Squawk Box', and `Last Man Standing', but the human clarified .. & Many summaries show cases like this where the assistant's initial answer is portrayed to be the wrong answer, because the summarizer put the human's requirements (for the query) before they actually occur in the dialogue. \\ \bottomrule
    \end{tabular}
    }
    \caption{Examples showing the common positions of errors in dialogue summaries. The first example is a typical case of how the summary contains \extcont at the end. The second example represents typical \turnseq errors in a summary, as also demonstrated in the example in Figure \ref{fig:new-intro-fig}.}
    \label{tab:pos-of-error}
\end{table*}

\section{Notation} \label{app:notation}
% \sahana{move the long notation and bacc to appendix, and make it smaller here, make equation in-line and longer in appendix}
\textbf{Input:} We have a dialogue $\textbf{C}$ composed of turns $\{t_1,t_2,\dots,t_n\}$, and its summary $\textbf{S}$ composed of sentences $\{s_1,s_2,\dots,s_n\}$. For a given error $\textbf{E}$, we have gold label $G_E \in \text{\{YES, NO\}}$ that indicates if the summary exhibits error $\textbf{E}$, and gold sentence set $G_s \subseteq \{s_1,s_2,\dots,s_n\}$ that indicates the summary sentences (if any) which exhibit the error. \textbf{Note:} If the error $\textbf{E}$ is \missturn, we would have $G_t \subseteq \{t_1,t_2,\dots,t_n\}$ instead of $G_s$, since the LLM's goal would be to identify which turns were missed out in the summary. To simplify writing, we only mention the sentence set in the notations below; however, we replace them with the turn set whenever we experiment with this error. 

\textbf{Task:} The task is to measure an LLM $\theta$'s ability to predict if the summary exhibits error $\textbf{E}$; the LLM has to give the predicted label $P_E\in \text{\{YES,NO\}}$ and the predicted sentence set $P_s \subseteq \{s_1,s_2,\dots,s_n\}$. We use a few-shot prompt $FP_E$ (that contains the definition for $\textbf{E}$ and example demonstrations) for the same. 

\textbf{Method:} The LLM takes $\textbf{C, S}$ and $FP_E$ as input. For each sentence $s_i \in \textbf{S}$, the LLM $\theta$ predicts label $p_i \in \text{\{YES,NO\}}$ that states whether $s_i$ exhibits error $\textbf{E}$ or not. We use these sentence-level predicted labels to get the overall predicted label and sentence set as follows: 
% We design few-shot prompt $FP_E$ that contains the definition of $E$ and few demonstrations for the same. We want LLM $J$ to use the prompt $FP_E$ to annotate whether each sentence $s_i \in S$ exhibits the error or not ($p_i \in \text{\{YES,NO\}}$), and also a short reasoning for the same. 
\begin{align}
    &\{p_1, p_2, \dots, p_n\} = \theta(FP_E, C, S)  \\
    &\mathbf{P_E} = 
    \begin{cases}
    \text{YES}, & \text{if any } p_i = \text{YES},\\
    \text{NO},  & \text{otherwise}.
    \end{cases} \\
    &\mathbf{P_s} = \{\, s_i \mid p_i = \text{YES} \,\}
\end{align}
% \noindent Note: for the \missturn error, we do each \textit{turn in the conversation} instead of each sentence in the summary. 

\textbf{Evaluation.} We calculate and report \textit{balanced accuracy}\footnote{\texttt{sklearn.metrics.balanced$\_$accuracy$\_$score}} for the overall labels $(P_E, G_E)$ as well as the sentence-level labels (YES/NO labels for $P_s, G_s$) for all models and errors. We term these \textbf{BAcc.} and \textbf{S-BAcc.} respectively. Note that we skip \spkbias (which does not occur in \dataset) and \missconv (as explained above, this error is too detailed, and we test our judges only on the higher-level incompleteness error, \missturn; we leave work on lower levels of incompleteness to future work). Note that for the broad, hallucination error, the gold labels and gold sentence sets are obtained by aggregating the gold labels and sentence sets for all fine-grained errors that are defined under the hallucination error. 
% The model: We design few-shot prompts to detect each error in \taxonomy, and test them with the following LLMs from various model families: \gptfive, \claudefull, \llamafull, \qwenfullone and \qwenfulltwo (full model details in Appendix \ref{xx}).
% \sahana{todo: discussion on results}
% API based model, open source model, also including rubric based reward model, different ways to format the prompt
% \sahana{todo: multi-judge votes, vectara}
Also, since we measure performance using balanced accuracy, 50\% is the accuracy for random guessing, and the higher the balanced accuracy gets, the better the model is at identifying the correct error labels.  

\section{Reproducibility}
\textbf{Data usage and license.}  We use \textsc{Anthropic-Test}\footnote{\texttt{Anthropic/hh-rlhf}, `helpful-base' on HuggingFace.}  \cite{bai2022training} as our base dataset. \textsc{Anthropic-Test} is open-sourced on huggingface, and is licensed to be freely used and distributed. \textsc{Anthropic-Test} includes human-AI assistant conversations, however the dataset has \textit{already been anonymized}; we do not take any further steps to detoxify the data, since we use the `helpful' subset of data.

\textbf{Models usage and license.} We use publicly available models \gptfive, \claude, \llamafull, \qwenfull, \llamasumm, and we perform only inference and no training.  We use \gptfive and \claude as paid API services. We use \llamafull, \llamasumm and \qwenfull from huggingface; these are open-source models, and we had to either request access and/or provide contact and usage information, and accept terms and conditions to access the models. All models were utilized consistently with their intended usage.

\textbf{Prompts and hyperparameters.} We release all prompts we use for summary generation as well as error detection in Appendix \ref{app:all-prompts}. All code and data will be released soon. We disclose all generation parameters in Section \ref{sec:datasets} (summary generation) and in Section \ref{sec:judge-exp} (error detection by LLMs); we further note here that we fixed the error detection temperature, top-p, etc. after testing with a range of values (\texttt{temperature} $\in \{0.6, 0.7, 0.9, 1.0\}$, \texttt{top-p} $\in \{0.7, 0.9, 0.95\}$, \texttt{verbosity, reasoning} for \gptfive $\in \{\text{low,medium,high}\}$. 

\textbf{Descriptive statistics about model runs.} As we mention in Section \ref{sec:judge-exp}, we sample 5 times for every evaluation instance, for every error. We take the aggregate prediction as our final prediction. 

\textbf{Disclosure of AI usage in plotting graphs.} We used \href{https://gemini.google.com/app}{Gemini}'s assistance in writing the python \texttt{matplotlib} code to plot graphs in Figures \ref{fig:freq_stats} and \ref{fig:start-mid-end}. We do not use AI otherwise for assistance in coding or in writing.

\end{document}